\pdfoutput=1

\documentclass[11pt]{article}

\usepackage[]{acl}

\usepackage{times}
\usepackage{latexsym}

\usepackage[T1]{fontenc}

\usepackage[utf8]{inputenc}

\usepackage{microtype}

\usepackage{graphicx} 
\usepackage{booktabs}

\usepackage{amsmath}
\usepackage{multirow} 
\usepackage{tabularx}

\title{Mind the Knowledge Gap:\\A Survey of Knowledge-enhanced Dialogue Systems}

\author{$^\star$Sagi Shaier, $^\dag$Lawrence Hunter, $^\star$Katharina Kann \\ $^\star$University of Colorado Boulder \\ $^\dag$University of Colorado Denver\\
$^\star$\{sagi.shaier, katharina.kann\}@colorado.edu\\
$^\dag$larry.hunter@cuanschutz.edu
}

\begin{document}
\maketitle
\begin{abstract}
Many dialogue systems (DSs) lack characteristics humans have, such as emotion perception, factuality, and informativeness. Enhancing DSs with knowledge alleviates this problem, but, as many ways of doing so exist, keeping track of all proposed methods is difficult. Here, we present the first survey of knowledge-enhanced DSs. We define three categories of systems -- internal, external, and hybrid -- based on the knowledge they use. We survey the motivation for enhancing DSs with knowledge, used datasets, and methods for knowledge search, knowledge encoding, and knowledge incorporation. Finally, we propose 
how to improve existing systems based on theories from linguistics and cognitive science. 

\end{abstract}

\section{Introduction}
Dialogue systems (DSs) are designed to communicate with humans, either for entertainment or for task completion.
While the movement from rule-based systems to machine learning models has improved the diversity of the generated responses, they still lack many attributes human language exhibits, such as emotion perception, factuality, and informativeness. This may in turn result in user frustration, or even harm, if the requested information returned by the DS is fictitious.

Previous work has found that incorporating knowledge (e.g., as knowledge graphs) into DSs alleviates many such problems, e.g., by boosting informativeness
\cite{DBLP:journals/corr/abs-2010-10150}, emotional intelligence \cite{Zhou_Huang_Zhang_Zhu_Liu_2018, Liang_Meng_Zhang_Chen_Xu_Zhou_2021}, or factuality \cite{dziri-etal-2021-neural}. However, an abundance of knowledge forms exist, many of considerable size. Hence, numerous methods for finding, encoding, and incorporating the relevant information from the knowledge source into the dialogue have been proposed. To the best of our knowledge, we provide the first survey to overview such works. 

\begin{figure*}[t]
    \centering
    \includegraphics[width=1.5\columnwidth]{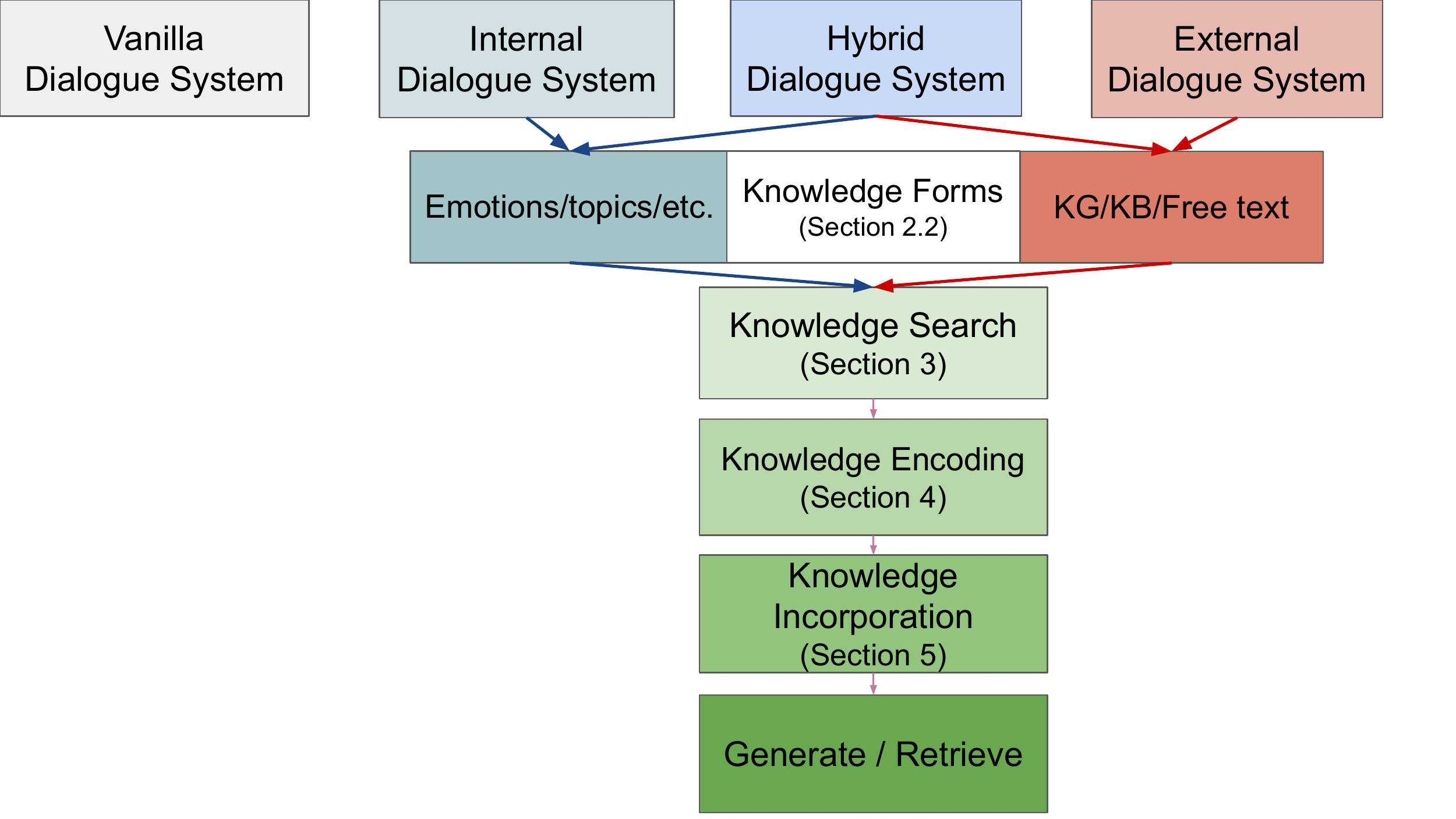}
    \caption{The steps conducted by dialogue systems. If knowledge is used, its type (knowledge forms, Section \ref{knowledge_forms_subsection}) needs to be considered. This will determine if the system is internal, external, or hybrid. Next, we search for the relevant knowledge in the often massive knowledge database (knowledge search, Section \ref{search}). Then, encoding of the relevant knowledge is needed (knowledge encoding, Section \ref{encoding}) followed by incorporating the output encoding into the encoded dialogue history (knowledge incorporation, Section \ref{incorporation}). Lastly and similarly to the vanilla systems that do not use knowledge, the system generate or retrieve a response based on the encoded representation.}
    \label{prelim_fig}
\end{figure*}

More specifically, we group knowledge-enhanced DSs based on the type of knowledge they use: \textbf{internal}, \textbf{external}, and \textbf{hybrid}. We discuss motivations for enhancing DSs with knowledge, typically used datasets, and, based on how humans enhance dialogues with knowledge, present existing methods for \textbf{knowledge search}, \textbf{knowledge encoding}, and \textbf{knowledge incorporation}. Lastly, using theories derived from the human memory system, we propose possible avenues to improve knowledge-enhanced DSs.


\section{Background}

\subsection{Knowledge}
\label{background_knowledge_subsection}
In the context of this paper, we define knowledge as any information that goes beyond the pure surface form of the utterances in the dialogue history. For example, if the dialogue history is 

\begin{center}
1) Good morning, how can I help you? \\
2) OMG I already told you!!! Where is the nearest bank??
\end{center}

The pure surface form is simply the characters or words (i.e., "Good", "morning", ..., "??"). In comparison, knowledge is information in various forms such as knowledge graph (KG) triples about the nearest bank (e.g., [Bank\_of\_America, distance, 0.1]), the user's emotional category (e.g., angry), an abstract meaning representation (AMR) describing the semantic association between words (e.g., "where" $\rightarrow$ question), or free text describing relevant information (e.g., "there are several banks in Colorado..."). 


\subsection{Knowledge Forms}
\label{knowledge_forms_subsection}
Surveying existing work, we identify three broad categories of knowledge (according to our definition) that DS have been enhanced with, and 
structure the survey according to these categories. That is, the existing work can be grouped into the \textbf{category} of knowledge that they use (i.e., internal, external, or hybrid), where each knowledge category is based on the \textbf{location} from which the knowledge was taken (e.g., KGs, Wikipedia articles, the utterance itself).

\textbf{Internal knowledge} is implicit information that is extracted from the dialogue and made explicit. In simple words, this is information that can be seen as "between the lines" of the dialogue. It can be structured, such as AMR graphs or dependency trees (e.g., "where" $\rightarrow$ question), or unstructured, such as conversation topics (e.g., direction assistance), or the speaker's emotional state (e.g., angry). While these may require predefined concepts (e.g., emotional categories, linguistic structures, topics), the actual information is found \textbf{within} the dialogue.

In contrast, \textbf{external knowledge} is information that is not extracted from the dialogue and exists in an \textbf{outside} source. It can be structured, such as KGs and knowledge bases (KBs), or unstructured, such as free text, and take the form of triples, table entries, or strings. 


The third, \textbf{hybrid knowledge}, is a combination of internal and external knowledge, such as emotion categories and KGs, topic representations and KBs, or intent vectors, KGs, and free text. 

We call enhanced DS that use internal or external knowledge \textbf{internal} and \textbf{external} approaches, respectively. Those that use a combination of both are called \textbf{hybrid} approaches.

\subsection{Dialogue Systems}
\label{DS}

There are two classes of DS: \textit{task-oriented}, which are designed to assist with task completion, and \textit{chit-chat}, designed mainly for entertainment. Each class may be based on response retrieval or response generation.
In the retrieval method, the DS is given a corpus of predefined responses and is tasked with selecting the most appropriate \cite{tao2021a}, while in the generative method the DS generates a response token by token \cite{DBLP:journals/corr/abs-1711-01731}. 

The simplest approach for retrieval systems is using 2 encoders: one to encode each response and a second to encode the dialogue, followed by a similarity function between each encoded response and the dialogue. For generative systems, one may use an encoder-decoder setup, where the decoder is tasked to generate a token at a time conditioning on the encoded dialogue and the previously generated tokens. For more information about DS we refer the interested reader to \citet{Jurafsky2008-zb} or  \citet{McTear2021}.

While some DS use knowledge more often, here, we will not focus on the distinction between task and non-task oriented systems or open and closed domain. Instead we will focus on knowledge and how different knowledge sources have been used.

\subsection{Knowledge Enhancement Process}
The breakdown we present can be seen in Figure \ref{prelim_fig}. The top row shows the three knowledge-enhancement approaches discussed before (internal, external, and hybrid), in addition to the \textbf{vanilla} approach, which does not use knowledge.

Given our example dialogue history in Section \ref{background_knowledge_subsection}, the vanilla approach encodes the utterances followed by the steps described in Section \ref{DS} regarding generation or retrieval systems. However, if knowledge is used, the type of knowledge must first be chosen (Section \ref{knowledge_forms_subsection}). Then, as the knowledge source may contain million data-points (e.g., KG triples), hindering the use of its entirety simultaneously, we use a method to find relevant parts. For example, by checking whether each triple has the word "bank" and retrieving some number of them. We discuss additional \textbf{knowledge search} methods in Section \ref{search}.


Once we find the relevant information in our knowledge source (e.g., 5 triples representing the nearest banks) we need to encode it, as it is often presented in a different form than the utterance (e.g., KG triples versus free text). For example, by encoding both the dialogue text and the triples into similar size vectors. We discuss more \textbf{knowledge encoding} methods in Section \ref{encoding}. 

Once both the knowledge and dialogue history are encoded, we need to incorporate the knowledge representations into the encoded utterance. For example, by summing both vector representations together. We discuss \textbf{knowledge incorporation} methods further in Section \ref{incorporation}. 

Lastly and similarly to the final stage in the vanilla approach, we generate or retrieve a response given our final representation. 

\subsection{Motivation for Enhancing Dialogue Systems With Knowledge}
\label{knowledge_motivation_cog}
While the vanilla approach assumes that  all the information to generate or retrieve a response is provided in the dialogue, this is often not the case. For example, some suggest that DS have limited ability to represent core semantics, such as ignoring relevant entities \cite{bai-etal-2021-semantic, Chen2016KnowledgeAA, su-etal-2018-natural}. DS may also have difficulties with perception and expressions of emotions \cite{Liang_Meng_Zhang_Chen_Xu_Zhou_2021, Zhou_Huang_Zhang_Zhu_Liu_2018, pmlr-v158-compton21a} or with being specialized to certain users or domains \cite{10.1145/3308558.3313498, DBLP:journals/corr/abs-2004-03588, DBLP:journals/corr/abs-2005-10450, moghe-etal-2020-incorporating}. For example, in the medical domain DS must generate meaningful responses with the correct entities, which they often not \cite{DBLP:journals/corr/abs-2108-01266}. 

Another crucial piece that is often missing from DS responses is factuality \cite{DBLP:journals/corr/abs-2107-07566, DBLP:journals/corr/abs-1910-07834}, which hinders them from being used in domains where human lives are at stake. DS often also generate responses that are too short, dull, or uninformative \citet{DBLP:journals/corr/abs-2005-00613, yavuz-etal-2019-deepcopy, DBLP:journals/corr/abs-2010-10150, parthasarathi-pineau-2018-extending, DBLP:journals/corr/abs-1903-09813, moon-etal-2019-opendialkg, DBLP:journals/corr/abs-1902-04911, DBLP:conf/emnlp/YangZE20, DBLP:journals/corr/abs-2104-11454}, and can be improved on answering simple factoid QA \cite{yin-etal-2016-neural-generative} which can be seen as a single-turn dialogue, or their reasoning skills \cite{DBLP:journals/corr/abs-2001-10468}.

DS also often have difficulties with implicit knowledge perception or linguistic features \cite{DBLP:journals/corr/abs-2009-09708, zhou-etal-2021-think, kumar-etal-2020-amused, DBLP:journals/corr/abs-2110-07477}, responding to user requests that are beyond the scope of existing APIs \cite{jin-etal-2021-assistance}, dealing with out-of-vocabulary entities \cite{DBLP:journals/corr/abs-1709-04264}, and performing well given limited resources \cite{DBLP:conf/iclr/ZhaoWTX0020}. Such challenges have been tackled by enhancing DS with knowledge, as will be seen shortly. 

\section{Knowledge Search}
\label{search}
We define knowledge search as the methods for \textbf{finding and extracting} the relevant knowledge from the data source (e.g., KG). Although the internal approach is mostly based on extraction (e.g., extracting AMR graphs), the external approach could be using both. However, the search aspect is often more important, but difficult. The following two subsections describe the internal and external approaches taken for knowledge search, with the hybrid approaches split into their external and internal parts and written within.

\subsection{Internal Knowledge Search} 
The methods for internal knowledge search can be seen in Table \ref{Internal_Knowledge_Search_table}.
Many focus on extracting linguistic features from the dialogue, such as AMR graphs, dependency trees, or POS tags. These are usually done using an existing library (e.g., NLTK, spaCy). In comparison, for emotional information, speaking style, or user intent, existing work often train a system to classify or extract the knowledge. 
Additionally, for a given knowledge form, e.g., domains or entities, several possible search methods exist, such as string, n-gram matching, 
or a classifier that tags relevant words.

\begin{table}[t]
\centering
\tiny
\caption{Internal Knowledge Search Table}
\label{Internal_Knowledge_Search_table}
\begin{tabularx}{\linewidth}{|p{0.1\textwidth}|p{0.1\textwidth}|X|}
\hline
\textbf{Knowledge Form}           & \textbf{Name}                                                                           & \textbf{Method}              \\ \hline
AMR graph                         & \cite{bai-etal-2021-semantic, Chen2016KnowledgeAA}                     & AMR parser                   \\ \hline
Dependency tree &
  \cite{DBLP:conf/aaai/WangLBLHXY20, Chen2016KnowledgeAA, yang-etal-2020-graphdialog} &
  Dependency parser \\ \hline
Part-of-speech tagging &
  \cite{su-etal-2018-natural, DBLP:journals/corr/abs-2111-05204} &
  POS tagger \\ \hline
Emotions &
  \cite{Liang_Meng_Zhang_Chen_Xu_Zhou_2021, pmlr-v158-compton21a, Zhou_Huang_Zhang_Zhu_Liu_2018} &
  Emotion classifier \\ \hline
\multirow{3}{*}{Specific domains} & \cite{DBLP:journals/corr/abs-2104-11454}                               & Topic classifier             \\ \cline{2-3} 
                                  & \cite{chaudhuri-etal-2018-improving}                                   & String matching              \\ \cline{2-3} 
                                  & \multirow{2}{*}{\cite{DBLP:journals/corr/abs-2106-09174}}              & Domain classifier            \\ \cline{1-1} \cline{3-3} 
\multirow{2}{*}{Entities}         &                                                                                         & Fuzzy n-gram matching        \\ \cline{2-3} 
                                  & \cite{DBLP:journals/corr/abs-2108-01266, DBLP:conf/esws/ChaudhuriR021} & Entity classifier            \\ \hline
\multirow{3}{*}{User intent} &
  \cite{DBLP:journals/corr/abs-2108-11063} &
  Expression matching, classifier \\ \cline{2-3} 
                                  & \cite{DBLP:journals/corr/abs-2001-10468}                               & Intent classifier            \\ \cline{2-3} 
                                  & \cite{DBLP:journals/corr/abs-1709-04264}                               & \multirow{2}{*}{GRU encoder} \\ \cline{1-2}
Speaking style                    & \cite{DBLP:journals/corr/abs-2107-07771}                               &                              \\ \hline
\end{tabularx}
\end{table}

\subsection{External Knowledge Search}
The methods for external knowledge search can be seen in Table \ref{External_Knowledge_Search_table}. 
In contrast to the string methods which examine whether the dialogue and knowledge source share strings in a binary fashion, such as having an exact string match, or non-stopwords match, the similarity methods use a score between 0 and 1 to evaluate how similar the knowledge and dialogue are. 
Attention mechanisms also use a scoring approach, but solely use attention without a specialized model, normally on embedding vectors.
There are also methods that are highly specific to the type of knowledge, such as graph linking methods which links the entities found in dialogue with the corresponding graph entities, and query methods, which use a specialized syntax, such as SQL. Some methods can avoid the search entirely, for example, by using the knowledge as labels in response generation and by that "injecting" it into the parameters of the model, or by using the entire KB in response generation. 

\begin{table*}
\caption{External Knowledge Search Table}
\label{External_Knowledge_Search_table}
\centering
\tiny
\begin{tabularx}{\linewidth}{|p{0.1\textwidth}|p{0.18\textwidth}|p{0.5\textwidth}|p{0.11\textwidth}|}
\hline
\textbf{Method Type} &
  \textbf{Knowledge Form} &
  \textbf{Name} &
  \textbf{Method} \\ \hline
\multirow{8}{*}{Similarity} &
  \multirow{2}{*}{Knowledge graph} &
  \cite{DBLP:journals/corr/abs-1910-07834, zhou-etal-2021-think} &
  Cosine similarity \\ \cline{3-4} 
 &
   &
  \cite{DBLP:journals/corr/abs-2104-11454} &
  TF-IDF \\ \cline{2-4} 
 &
  Knowledge base &
  \cite{yin-etal-2016-neural-generative, DBLP:conf/aaai/WangLBLHXY20, ramadan-etal-2018-large} &
  Dot product / FC layer \\ \cline{2-4} 
 &
  \multirow{5}{*}{Free text} &
  \cite{cai-etal-2019-skeleton} &
  Jaccard distance \\ \cline{3-4} 
 &
   &
  \cite{DBLP:journals/corr/abs-1902-04911, wu-etal-2021-dialki, DBLP:conf/nlpcc/LiuZLR21} &
  Dot product \\ \cline{3-4} 
 &
   &
  \cite{jin-etal-2021-assistance, DBLP:journals/corr/abs-2102-02096, roller-etal-2021-recipes} &
  Transformer \\ \cline{3-4} 
 &
   &
  \cite{zhang-etal-2021-smoothing, yavuz-etal-2019-deepcopy, roller-etal-2021-recipes} &
  TF-IDF \\ \cline{3-4} 
 &
   &
  \cite{DBLP:journals/corr/abs-2010-10150, 10.1561/1500000019} &
  Cosine similarity \\ \hline
Graph linking &
  \multirow{3}{*}{Knowledge graph} &
  \cite{DBLP:journals/corr/abs-2110-07477, kumar-etal-2020-amused} &
  Entity linking \\ \cline{1-1} \cline{3-4} 
\multirow{5}{*}{String} &
   &
  \cite{DBLP:journals/corr/abs-2009-09708} &
  Non-stopword matching \\ \cline{3-4} 
 &
   &
  \cite{zhou-etal-2021-think} &
  Lemmatized words matching \\ \cline{2-4} 
 &
  Knowledge base &
  \cite{he-etal-2017-learning, wang-etal-2021-template, DBLP:journals/corr/abs-1709-04264, agarwal-etal-2018-knowledge} &
  Exact matching \\ \cline{2-4} 
 &
  \multirow{2}{*}{Free text} &
  \cite{DBLP:journals/corr/abs-2012-11937} &
  Exact matching \\ \cline{3-4} 
 &
   &
  \cite{DBLP:journals/corr/abs-2005-00613} &
  N-gram matching \\ \hline
\multirow{3}{*}{Query} &
  Knowledge graph &
  \cite{DBLP:conf/esws/ChaudhuriR021, DBLP:journals/corr/abs-2112-07924, DBLP:journals/corr/abs-2108-11063} &
  SPARQL \\ \cline{2-4} 
 &
  \multirow{2}{*}{Knowledge base} &
  \cite{DBLP:journals/corr/abs-1906-06788} &
  Symbolic query \\ \cline{3-4} 
 &
   &
  \cite{madotto-etal-2020-learning} &
  SQL, CYPHER \\ \hline
Memory networks &
  \multirow{2}{*}{Knowledge base} &
  \cite{DBLP:journals/corr/abs-1911-08522, qin-etal-2019-entity} &
  Memory networks \\ \cline{1-1} \cline{3-4} 
\multirow{3}{*}{Attention} &
   &
  \cite{DBLP:conf/emnlp/YangZE20, le-etal-2016-lstm, wen-etal-2018-sequence, eric-etal-2017-key} &
  Attention mechanism \\ \cline{2-4} 
 &
  Free text &
  \cite{moghe-etal-2020-incorporating, ma-etal-2020-compare, DBLP:journals/corr/abs-1906-06685, DBLP:conf/iclr/KimAK20, liu-etal-2021-three, yavuz-etal-2019-deepcopy, DBLP:conf/iclr/ZhaoWTX0020, zhao-etal-2020-knowledge-grounded} &
  Attention mechanism \\ \cline{2-4} 
 &
  Knowledge graph &
  \cite{moon-etal-2019-opendialkg} &
  Attention mechanism \\ \hline
\multirow{2}{*}{NA} &
  Knowledge base &
  \cite{DBLP:conf/emnlp/GouLLDS21} &
  Entire table \\ \cline{2-4} 
 &
  Free text &
  \cite{DBLP:journals/corr/abs-2105-06232, cui-etal-2021-knowledge, xu-etal-2021-k-plug, DBLP:journals/corr/abs-2005-10450} &
  Knowlede injection \\ \hline
Nearest neighbors &
  Free text &
  \cite{DBLP:journals/corr/abs-2107-07566, fan-etal-2021-augmenting} &
  Nearest-neighbors \\ \hline
\end{tabularx}
\end{table*}


\paragraph{Reflection}

The similarity methods, which in contrast to many other methods, such as using existing parsers or queries, are easily parallelizable and differentiable. However, they often require the system to be trained which is more time consuming but does not require experts to create rules.
And while the similarity methods can be parallalized, the string-based approach is simple and direct. 

Query methods are designed to be efficient, but require user knowledge and a database that is structured according to the syntax. Memory networks \cite{DBLP:journals/corr/SukhbaatarSWF15} have also been used, as they appear to be effective in incorporating KBs into neural models. Although, the more efficient methods are those that avoid the search entirely by injecting the knowledge. However, these are complex to develop and have several issues that will be discussed in the following sections.

In comparison to the amount of papers focusing on external knowledge there is much less work done on internal knowledge, even though that we identify 3 times as many internal knowledge forms used than external ones. This may be due to lack of internal datasets visibility, which we hope to alleviate in this survey (Section \ref{datasets}). 

The existing work highlight two main problems: 1) only locating the exact relevant knowledge, 2) doing so using methods that are suitable for real-time applications. For the first, considering the enormous amount of data, only selecting the exact relevant knowledge is impressive if not ambitious; especially given that such relevant knowledge may not appear in the same sentence, triple, knowledge source, or even exist in the data at all. However, many works assume that such relevant knowledge exist in a single location. Additionally, very few use entity linking on KGs, which are often encoded, rather than string based methods. This might imply that there is a gap between those who build such KGs and those who use them.

Many also use search methods that retrieve the top-K data which is problematic for four reasons. 1) this approach ignores the fact that some dialogues require more or less information. 2) a main reason for developing KGs/KBs triples is to reduce noise, but using all retrieved triples might introduce even more noise. 3) the approach assumes that the knowledge database is complete (i.e., the knowledge always exists). 4) the length of the concatenated retrieved knowledge might exceed the length limitation of language models (LM) and hence truncated and potentially lose the actual relevant knowledge. 

For the second problem, many use methods that are impractical for real-time applications, such as matching over all of the knowledge in the database. Notably, rarely any work mentions the time it takes for systems to work, which may significantly hinder their usage. For example, a DS that is nearly perfect but takes a month to respond will rarely be used. 

\paragraph{Cognitively-Inspired Future Work}
\label{search_cog}

Semantic memory networks and clusters theories, like those in \citet{Pereira2018-qr, Hills2012-pj, NIPS2012_14d9e800}, which follow the theory that memories are organized in a network structure, suggest that current knowledge search methods can be improved. For example, instead of searching through the entire knowledge source (e.g., by string matching over each triple), by structuring knowledge in such semantic network and clusters we can use random walks \cite{NIPS2012_14d9e800} 
or perform optimal foraging policy methods \cite{Hills2012-pj}. \citet{Brain_reflections} theory could also be used, by 
structuring the knowledge source in such way that it will show differences when different knowledge types are triggered

Alternatively and the quickest option, is to follow \citet{Buckner2001-up}'s theory about spontaneously activated memories, using models that avoid the search entirely. 
However, such methods have limitations, such as the inability to use newly-added knowledge without retraining.
Future work should examine methods to avoid searching over all knowledge data, as it significantly limits the real-time applications of the DS and hence its use case. Furthermore, similarly to the human brain where knowledge search must be done quickly for survival and effective communication, DS should also aim for such speed. Future work should emphasize time in future challenges and datasets as a metric for system performance. 


It is important to remember that knowledge search is still very much an open problem. After all, humans have not perfected the task either, which can be seen when attempting to retrieve known answers to questions unsuccessfully. However as stated in \cite{Pereira2018-qr}, humans have no direct access to their semantic knowledge network. That being said, DS do, which implies a potential to bypass human performance in such tasks.

\section{Knowledge Encoding}
\label{encoding}
As the dialogue and knowledge often appear in different forms (e.g., free text and KG), we need to encode them to similar form so they can be incorporated later. Here we present such methods, which take knowledge in various forms and encode it. We assume that the dialogue is encoded in one of many ways which can be found in one of the resources we referenced in Section \ref{DS}.

\subsection{Internal Knowledge Encoding}
The methods for internal knowledge encoding can be seen in Table \ref{Internal_Knowledge_Encoding_table}.
As dialogue has a time component to it, most work use models that can manage such information, such as RNNs, GRUs, and Transformers. While the latter has become the most popular in recent year for its long time dependencies. Trainable parameters are also used instead of a full model, such as representing the speakers using token embeddings and the emotions as trainable matrices.

\begin{table}[t]
\caption{Internal Knowledge Encoding Table}
\label{Internal_Knowledge_Encoding_table}
\centering
\tiny
\begin{tabularx}{\linewidth}{|p{0.15\textwidth}|p{0.1\textwidth}|p{0.15\textwidth}|}
\hline
\textbf{Knowledge Form}                & \textbf{Name}                                          & \textbf{Method}         \\ \hline
\multirow{2}{*}{AMR graph}             & \cite{bai-etal-2021-semantic}                             & Graph transformer       \\ \cline{2-3} 
                                      & \cite{Chen2016KnowledgeAA}                                & FC network / RNN / CNN  \\ \hline
\multirow{2}{*}{Dependency tree}       & \cite{DBLP:conf/aaai/WangLBLHXY20}                        & GRU                     \\ \cline{2-3} 
                                      & \cite{yang-etal-2020-graphdialog}                         & Graph encoder           \\ \hline
Part-of-speech tagging                 & \cite{su-etal-2018-natural}                               & NA: Knowledge injection \\ \hline
Noun phrases                           & \cite{DBLP:journals/corr/abs-2111-05204}                  & Transformer             \\ \hline
\multirow{3}{*}{Emotions} & \cite{Liang_Meng_Zhang_Chen_Xu_Zhou_2021}           & Trainable parameters    \\ \cline{2-3} 
                                      & \cite{Zhou_Huang_Zhang_Zhu_Liu_2018}                 & GRU                     \\ \cline{2-3} 
                                      & \cite{pmlr-v158-compton21a}                               & Transformer             \\ \hline
\multirow{3}{*}{Specific domains}      & \cite{DBLP:journals/corr/abs-2104-11454}                  & Transformer             \\ \cline{2-3} 
                                      & \cite{chaudhuri-etal-2018-improving}                      & GRU                     \\ \cline{2-3} 
                                      & \multirow{2}{*}{\cite{DBLP:journals/corr/abs-2106-09174}} & Transformer             \\ \cline{1-1} \cline{3-3} 
\multirow{2}{*}{Entities}              &                                                        & Transformer             \\ \cline{2-3} 
                                      & \cite{DBLP:journals/corr/abs-2108-01266}                  & Transformer                 \\ \hline
\multirow{2}{*}{User intent}           & \cite{DBLP:journals/corr/abs-2001-10468}                  & RNN                     \\ \cline{2-3} 
                                      & \cite{DBLP:journals/corr/abs-1709-04264}                  & RNN                     \\ \hline
Speaking style                         & \cite{DBLP:journals/corr/abs-2107-07771}                  & GRU                     \\ \hline
Speaker representation                 & \cite{DBLP:journals/corr/abs-2004-03588}                  & Trainable parameters    \\ \hline
\end{tabularx}
\end{table}

\subsection{External Knowledge Encoding}
The methods for internal knowledge encoding can be
seen in Table \ref{External_Knowledge_Encoding_table}.
Similarly to the internal encoding methods, many works use algorithms that take time into account, such as RNN, GRU, and Transformers. There also exist work on using graph algorithms to encode the KG triples, such as RGCNN, graph Laplacian, and graph transformers, and work that simply use trainable parameters, such as averaging the KG/KB entity embeddings. For free text, the bulk of the work use transformer-based models, while for KBs there is about equal number of works that use any method.

\begin{table*}[t]
\caption{External Knowledge Encoding Table. KG = knowledge graph, KB = knowledge base, FT = free text}
\label{External_Knowledge_Encoding_table}
\centering
\tiny
\begin{tabularx}{\linewidth}{|p{0.1\textwidth}|X|p{0.15\textwidth}|}
\hline
\textbf{Knowledge Form} &
  \textbf{Name} &
  \textbf{Method} \\ \hline
\multirow{6}{*}{Knowledge graph} &
  \cite{DBLP:journals/corr/abs-1910-07834, kumar-etal-2020-amused, DBLP:journals/corr/abs-2001-10468, DBLP:conf/emnlp/YangZE20} &
  Trainable parameters \\ \cline{2-3} 
 &
  \cite{dziri-etal-2021-neural, DBLP:journals/corr/abs-2104-11454, zhou-etal-2021-think} &
  Transformer \\ \cline{2-3} 
 &
  \cite{DBLP:journals/corr/abs-2110-07477} &
  RGCNN \\ \cline{2-3} 
 &
  \cite{DBLP:journals/corr/abs-2009-09708} &
  Graph transformer \\ \cline{2-3} 
 &
  \cite{moon-etal-2019-opendialkg} &
  NN \\ \cline{2-3} 
 &
  \cite{DBLP:conf/esws/ChaudhuriR021} &
  Graph Laplacian \\ \hline
\multirow{6}{*}{Free text} &
  \cite{DBLP:journals/corr/abs-1902-04911, DBLP:journals/corr/abs-1906-06685, yavuz-etal-2019-deepcopy, cai-etal-2019-skeleton, moghe-etal-2020-incorporating} &
  RNN \\ \cline{2-3} 
 &
  \cite{DBLP:conf/iclr/ZhaoWTX0020} &
  GRU \\ \cline{2-3} 
 &
  \cite{DBLP:journals/corr/abs-1903-09813} &
  CNN \\ \cline{2-3} 
 &
  \cite{zhang-etal-2021-smoothing, DBLP:conf/iclr/KimAK20, zhao-etal-2020-knowledge-grounded, wu-etal-2021-dialki, roller-etal-2021-recipes, DBLP:conf/nlpcc/LiuZLR21, DBLP:journals/corr/abs-2005-00613, DBLP:journals/corr/abs-2102-02096, DBLP:journals/corr/abs-2010-10150, hedayatnia-etal-2020-policy,  DBLP:journals/corr/abs-2012-11937, DBLP:journals/corr/abs-2107-07566, ma-etal-2020-compare, liu-etal-2021-three, fan-etal-2021-augmenting, DBLP:journals/corr/abs-2111-05204, jin-etal-2021-assistance} &
  Transformer \\ \cline{2-3} 
 &
  \cite{DBLP:journals/corr/abs-2111-05204, DBLP:journals/corr/abs-2105-06232, DBLP:journals/corr/abs-2005-10450, cui-etal-2021-knowledge, xu-etal-2021-k-plug} &
  Knowledge injection \\ \cline{2-3} 
 &
  \cite{parthasarathi-pineau-2018-extending} &
  Trainable parameters \\ \hline
\multirow{5}{*}{Knowledge base} &
  \cite{wang-etal-2021-template, DBLP:journals/corr/abs-1906-06788, DBLP:journals/corr/abs-1911-08522} &
  Memory network \\ \cline{2-3} 
 &
  \cite{yin-etal-2016-neural-generative, ramadan-etal-2018-large, DBLP:journals/corr/abs-1709-04264, he-etal-2017-learning, le-etal-2016-lstm} &
  RNN \\ \cline{2-3} 
 &
  \cite{DBLP:conf/aaai/WangLBLHXY20, agarwal-etal-2018-knowledge} &
  GRU \\ \cline{2-3} 
 &
  \cite{wen-etal-2018-sequence, eric-etal-2017-key, qin-etal-2019-entity} &
  Trainable parameters \\ \cline{2-3} 
 &
  \cite{DBLP:conf/emnlp/GouLLDS21, DBLP:journals/corr/abs-2112-07924, madotto-etal-2020-learning} &
  Transformer \\ \hline
\end{tabularx}
\end{table*}

 \subsection{Discussion \& Thoughts}

\paragraph{Reflection}
\label{encoding_reflection}
Methods for encoding knowledge are typically based on either getting a vector representation, or by representing structured knowledge as text. To obtain a vector representation, some use trainable embeddings to represent the knowledge, for example by representing each KG triple or KB entries as an average of the subject and relation or cells embeddings. This method is simplistic, as it does not require a model. However, such approach might not have the same encoding capabilities of transformer-based models. Others choose to use a model, for example by representing the KG triples, free text words, or KB cells as tokens and sending these to a transformer-based model. 

Knowledge encoding can also be categorized into three main general approaches: 1) combining the knowledge and dialogue (e.g., into one sequence) followed by a model for encoding, 2) encoding the dialogue and each relevant knowledge independently, 3) injecting the knowledge into the parameters of the model, for example by using the knowledge as a target for the model.

Each of the three has benefits and drawbacks. The first approach is rather simplistic, as the encoder has only one representation to encode. However, if such representation does not separate the knowledge and dialogue sufficiently, the knowledge may be lost in the output representation. The second is slightly more complex, as it potentially requires multiple encoders for the dialogue and each knowledge form. However, it allows for more freedom in the next step (i.e., incorporation). That is, as the knowledge and dialogue representations are separated, the model does not need to decipher which part of the representation contributes to the knowledge. The third is the most alluring as it avoids the search and encoding portions, but has several problems. As knowledge is dynamic and constantly changing, these methods require retraining models to encode the new knowledge. Additionally, there is a question of whether such methods can truly encode knowledge that rarely appear in the training data. 

While it is difficult converting free text into KGs, doing so for KBs is rather straight forward. Some works have shown that converting KBs to KGs have benefits \cite{he-etal-2017-learning, DBLP:conf/emnlp/YangZE20}, specifically, better representation of entities is possible as relational information between entities exist. That being said, not many use graph-based methods directly on KGs. 

Lastly, there is a question of whether a single vector representation, which most works use to encode the knowledge, is sufficient to encode all of the relevant information. This is apparent most often in free text, where the long documents may contain numerous relevant facts. 

\paragraph{Cognitively-Inspired Future Work}
While it is yet unclear how humans convert neuron-activity directly into words, there is evidence that specific neuronal activity corresponds to specific words. Cognitively, this section can be seen as “learning”, where memories, or knowledge, are encoded in the brain. Word embeddings can be seen as such encoding, specifically, a representation of the semantic networks discussed in Section \ref{knowledge_motivation_cog}. While most work use one embedding to represent the vocabulary, \citet{ross2010psychology}, \citet{Reddy1993MetaphorAT}, and \citet{Conversational_Evidence_for_Rethinking_Meaning} argue that words do not contain their meaning. By this argument, different speakers may have different meanings for a word. Hence, different speakers require different word embeddings to truly represent that humans have different memories and knowledge, and view the world differently. \citet{ross2010psychology} also argue that while speakers have their own representations of concept meanings, these meanings are fluid. That is, humans tailor their language to whom they talk to, and the listener interprets that language based on their knowledge of the speaker. This suggests that current fixed embeddings methods might not be sufficient, and a dynamic representation can be beneficial. 

Another important aspect of human knowledge is that it is dynamic. Our view of the world, memories, and knowledge are constantly changing
However, current systems largely assume a fixed knowledge source. This can be problematic when trained DS memorize and generate previously seen data rather than using the new data, or when the systems themselves do not have the capability of incorporating novel information without changing the architecture or retraining. Future work should examine methods of handling such dynamic knowledge source.

Lastly, according to the interactive alignment model \cite{pickering_garrod_2004}, language processing in monologues is different from that in dialogues. This suggests that current methods of pretraining large LM on data, such as Wikipedia text, may not represent how humans learn language, and hence develop insufficient world models and achieve lower performance. A potential avenue to explore is comparing the performance of such systems on monologues and dialogues, in addition to developing more such rich dialogue datasets. 

\section{Knowledge Incorporation}
\label{incorporation}
So far we have found the relevant knowledge to be used and encoded it. Now we need to incorporate it into the encoded dialogue history. Meaning, we need to combine the two encoded representations so our system can decide on an appropriate response. As the knowledge is encoded, we do not need to differentiate between the internal or external types. Hence, all of the incorporation methods can be seen in Table \ref{Knowledge_incorporation_table}, which has 6 different such methods.

\begin{table*}[]
\caption{Knowledge Incorporation Table}
\label{Knowledge_incorporation_table}
\centering
\tiny
\begin{tabularx}{\linewidth}{|p{0.2\textwidth}|X|}
\hline
\textbf{Method}                 & \textbf{Name}                                                                    \\ \hline
Aggregation followed by blending & \cite{bai-etal-2021-semantic, pmlr-v158-compton21a,  jin-etal-2021-assistance, DBLP:journals/corr/abs-2111-05204, Chen2016KnowledgeAA,  DBLP:conf/nips/LiX0ZZT20,DBLP:journals/corr/abs-2012-11937, hedayatnia-etal-2020-policy,DBLP:journals/corr/abs-2010-10150,DBLP:journals/corr/abs-2102-02096, DBLP:journals/corr/abs-2005-00613,DBLP:conf/nlpcc/LiuZLR21, roller-etal-2021-recipes,wu-etal-2021-dialki,DBLP:conf/iclr/KimAK20,zhao-etal-2020-knowledge-grounded, zhang-etal-2021-smoothing, yin-etal-2016-neural-generative, DBLP:journals/corr/abs-2004-03588,  Liang_Meng_Zhang_Chen_Xu_Zhou_2021,  DBLP:journals/corr/abs-2107-07771,  chaudhuri-etal-2018-improving,  dziri-etal-2021-neural, kumar-etal-2020-amused,  zhou-etal-2021-think,agarwal-etal-2018-knowledge, wen-etal-2018-sequence,DBLP:journals/corr/abs-2112-07924} \\ \hline
Attention mechanism             & \cite{bai-etal-2021-semantic, DBLP:conf/emnlp/YangZE20,DBLP:journals/corr/abs-2108-01266, moon-etal-2019-opendialkg, eric-etal-2017-key,moghe-etal-2020-incorporating,liu-etal-2021-three,DBLP:journals/corr/abs-1902-04911,DBLP:journals/corr/abs-1903-09813,ma-etal-2020-compare}                                                                                                                                                                                                                                                                                                                                                                                                                                                                                                                                       \\ \hline
Selection mechanism             & \cite{DBLP:journals/corr/abs-1910-07834, DBLP:journals/corr/abs-2110-07477,DBLP:conf/emnlp/GouLLDS21,qin-etal-2019-entity,DBLP:conf/iclr/ZhaoWTX0020,DBLP:journals/corr/abs-1709-04264}                                                                                                                                                                                                                                                                                                                                                                                                                                                                                                                                                                                                                                             \\ \hline
Knowledge injection             & \cite{DBLP:journals/corr/abs-2005-10450, su-etal-2018-natural,madotto-etal-2020-learning,DBLP:journals/corr/abs-2105-06232,cui-etal-2021-knowledge,xu-etal-2021-k-plug}                                                                                                                                                                                                                                                                                                                                                                                                                                                                                                                                                              \\ \hline
Graph creation                  & \cite{DBLP:journals/corr/abs-2009-09708,he-etal-2017-learning}                                                                                                                                                                                                                                                                                                                                                                                                                                                                                                                                                                                                                                                                                                                                                                      \\ \hline
Memory networks                 & \cite{ramadan-etal-2018-large, DBLP:journals/corr/abs-1906-06788,DBLP:journals/corr/abs-1911-08522,wang-etal-2021-template}                                                                                                                                                                                                                                                                                                                                                                                                                                                                                                                                                                                                                                                                                                         \\ \hline
                                                     
\end{tabularx}
\end{table*}

\subsection{Discussion \& Thoughts}
\paragraph{Reflection}

The main incorporation methods can be seen as: 1) aggregation of the representations of the dialogue and knowledge, followed by a blending mechanism. 2) attention over the representations. 3) a mechanism which selects which representation to use. 4) knowledge injection.

The bulk of the work has aggregated the encoded knowledge and dialogue (e.g., by summing vectors or concatenating tokens) followed by a a blending mechanism, such as a whole model, trainable parameters, or some nonlinear function. The blending mechanism is crucial, as if it does not combine the knowledge and dialogue well, the system may not understand what representation it should focus on. Unsurprisingly, transformer-based models appear in the highest frequency, likely for its great success in recent years. However, they often require long training time and a lot of data, which is often unavailable, specifically for unique domains, facts, or novel knowledge forms.

While many models have attention mechanism included in their architecture, many have also use simpler attention mechanisms to attend over the knowledge and dialogue representations. These require much less computational power and training time, but may have lower capabilities. Some also use selection mechanisms which select which representation to use in order the generate or retrieve an appropriate response. For example, by using the copy mechanism, which allows the incorporation of tokens directly from the knowledge source. This is especially useful when the knowledge has vocabulary that is not present in the training data, in addition to solving the dynamic-knowledge problem of knowledge-injection discussed in Section \ref{encoding_reflection}.

Knowledge injection methods also have several problems as mentioned in Section \ref{encoding_reflection}. Additionally, these approaches are form-specific, such as having unique losses for domains \cite{DBLP:journals/corr/abs-2005-10450} or POS tags \cite{su-etal-2018-natural}. This is unsustainable as DS should be able to incorporate multiple forms simultaneously. Simpler and less popular methods also exist, such as combining the dialogue and knowledge representations into one graph, or using memory networks most often on a KB.

\paragraph{Cognitively-Inspired Future Work}

The bulk of the work focus on a rather straightforward method of aggregating the knowledge and dialogue representations followed by a blending mechanism. However, 
the human language has a an extremely complex structure that may require many different rules. The study of pragmatics \cite{MEY200651} for example, focuses on how context contributes to language utilization, and encompasses several phenomena; implicature is when utterances contain implied information, while the relevance theory \cite{Sperber1995}, states that every utterance contains relevant, worth listening information. Knowledge-enhanced DS are taking the right steps towards such studies by explicitly incorporating information that is hidden but implied, in addition to improving the response's relevancy and quality. However, current DS incorporation methods are solely based on the encoded dialogue and very limited knowledge forms at a time. In comparison, humans use many context features which allow them to incorporate knowledge differently into the dialogue, such as explaining the same concept to a child versus an adult, and responding differently to different people based on their knowledge of the listener \cite{ross2010psychology}. 

In order for DS to successfully mimic human communication, future work should emphasize fluidity in their response and better models of the listeners, such as various characteristics and likes. For example, by taking attributes that are rarely used by DS these days, such as the users age, emotional status, or even time of the day, as we alternate our wordings based on listeners' maturity and generate shorter responses when we are busy. 

To communicate successfully, humans use a great amount of contexts that are based on the environment, listeners, and themselves, while DS use hardly any in comparison. Hence, a considerable potential for improvement exist.

\section{Datasets}
\label{datasets}
Table \ref{datasets_table} presents a comprehensive list of the datasets used in the reviewed work. It is split into the dataset name and its description.

\begin{table*}[]
\caption{Datasets Table}
\label{datasets_table}
\centering
\tiny
\begin{tabularx}{\linewidth}{|X|X|}
\hline
\textbf{Dataset Name}                                                                                   & \textbf{Description}                                                                                                               \\ \hline
Topical-Chat (TC) \cite{DBLP:conf/interspeech/GopalakrishnanH19}                                           & A knowledge-grounded conversations where the knowledge is on 8 topics                                                        \\ \hline
CMU Document Grounded Conversations \cite{DBLP:conf/emnlp/ZhouPB18}                                        & A knowledge-grounded conversations where the knowledge is about specific Wikipedia articles about popular movies                   \\ \hline
CMU Movie Summary \cite{DBLP:conf/acl/BammanOS13}                                                          & Summaries of movie plots extracted from Wikipedia with aligned metadata extracted from Freebase                                    \\ \hline
Reddit Conversation Corpus \cite{DBLP:journals/corr/abs-1811-01063}                                        & Dialogues extracted from Reddit, where each is composed of 3 turn exchanges                                                    \\ \hline
Reddit dataset \cite{DBLP:conf/emnlp/MazareHRB18}                                                          & Dialogues that are based on personas extracted from Reddit                                                                         \\ \hline
Reddit discussions (The Pushshift Reddit Dataset) \cite{DBLP:conf/icwsm/BaumgartnerZKSB20}                 & Submissions and comments posted on subreddits                                                                                      \\ \hline
Reddit comments \cite{boyd-etal-2020-large}                                                                & Conversations extracted from Reddit comments                                                                                       \\ \hline
Grounded Reddit conversation \cite{DBLP:conf/acl/QinGBLGDCG19}                                             & Conversations extracted from Reddit, which are linked to documents discussed in the conversations                                  \\ \hline
Soccer dialogues \cite{DBLP:conf/semweb/ChaudhuriRJ019}                                                    & Soccer dialogues along with a knowledge graph for each team                                                                        \\ \hline
In-car dialogue dataset \cite{DBLP:conf/sigdial/EricKCM17}                                                 & KB-grounded dialogues which span 3 distinct tasks in the in-car personal assistant space                                           \\ \hline
ConvAI2 dataset \cite{DBLP:journals/corr/abs-1902-00098}                                                   & Conversational dataset based on the PERSONA-CHAT dataset                                                                           \\ \hline
LIGHT dataset \cite{shuster-etal-2021-dialogue}                                                            & Episodes of character interactions in a text adventure game                                                                        \\ \hline
LightWild \cite{shuster-etal-2021-dialogue}                                                                & Episodes in a role playing game where players converse with learning agents                                                        \\ \hline
LightQA \cite{DBLP:journals/corr/abs-2111-05204}                                                           & A derived version of LightWild, ending on a question about the episode                                                             \\ \hline
DailyDialog \cite{DBLP:conf/ijcnlp/LiSSLCN17}                                                              & Multi-turn dialogues with human-written conversations                                                            \\ \hline
Empathetic Dialogues \cite{DBLP:conf/acl/RashkinSLB19}                                                     & Conversations that are grounded in emotional situations                                                                            \\ \hline
Blended Skill Talk \cite{DBLP:conf/acl/SmithWSWB20}                                                        & English conversations aimed at testing several skills, such as being engage, empathetic, and knowledgeable                                                               \\ \hline
OpenQA-NQ \cite{lee-etal-2019-latent}                                                                      & Google queries paired with short answers extracted from Wikipedia                                                                  \\ \hline
Multimodal EmotionLines Dataset \cite{DBLP:conf/acl/PoriaHMNCM19}                                          & Extension of EmotionLines dataset                                                                                                  \\ \hline
Interactive Emotional Dyadic Motion Capture Database \cite{DBLP:journals/lre/BussoBLKMKCLN08}              & Emotion-segmented videos of dyadic conversations                                                                                   \\ \hline
EmoryNLP \cite{DBLP:conf/aaai/ZahiriC18}                                                                   & Episodes, scenes, and emotional-grounded utterances                                                                                \\ \hline
EmpatheticDialogues \cite{rashkin-etal-2019-towards}                                                       & Emotional-grounded conversations                                                                                                   \\ \hline
MuTual \cite{cui-etal-2020-mutual}                                                                         & Multi-Turn dialogue reasoning dataset based on English listening comprehension exams taken by Chinese students                                \\ \hline
Crowd-sourced SocialIQA-prompted \cite{DBLP:journals/corr/abs-2109-06427}                                  & Dialogues that exhibit social commonsense in an interactive setting                                                           \\ \hline
Wizard of Wikipedia \cite{DBLP:conf/iclr/DinanRSFAW19}                                                     & Wikipedia-grounded Conversations with many discussions topics                                                                      \\ \hline
Wizard-of-Oz (WOZ) 2.0 \cite{DBLP:journals/corr/MrksicSWTY16}                                              & Expansion of the WOZ dataset                                                                                                       \\ \hline
MultiDomain Wizard-of-Oz (MultiWOZ) \cite{budzianowski-etal-2018-multiwoz}                                 & Annotated human-human written conversations spanning 8 domains                                                                     \\ \hline
MultiWOZ 2.1, 2.2 \cite{DBLP:conf/lrec/EricGPSAGKGKH20}, \cite{zang-etal-2020-multiwoz}                       & Extension of MultiWOZ                                                                                                              \\ \hline
OR-ShARC \cite{DBLP:journals/corr/abs-2102-08633}                                                          & Conversational machine reading dataset where the gold rule text for each sample is removed and used as a KB                  \\ \hline
Alexa Prize \cite{DBLP:journals/corr/abs-1812-10757}                                                       & Spoken conversations which are also not task-restricted, open-ended, topical, involve opinions, and are conducted with real users \\ \hline
Holl-E \cite{DBLP:conf/emnlp/MogheABK18}                                                                   & Movie conversations where each response is generated by copying and/or modifying sentences from unstructured background knowledge          \\ \hline
Dialog bAbI \cite{DBLP:journals/corr/BordesW16}                                                            & Noise-free simulated dialogues                                                                                                     \\ \hline
mDSTC2 dataset \cite{henderson-etal-2014-second}                                                           & Modified version of the DSTC2 dataset                                                                                              \\ \hline
DialogRE \cite{yu-etal-2020-dialogue}                                                                      & Relations-annotated dialogues originating from the complete transcripts of Friends                                                 \\ \hline
PersonaChat \cite{DBLP:conf/acl/KielaWZDUS18}                                                              & Multi-turn dialogues conditioned on personas                                                                                       \\ \hline
OpenSubtitles \cite{DBLP:conf/lrec/LisonT16}                                                               & Multilingual parallel corpora from a database of movies and TV subtitles                                                           \\ \hline
DSTC7-Track1 \cite{DBLP:journals/corr/abs-1901-03461}                                                      & Partial conversations which requires users to select the correct next utterances from a set of candidates      \\ \hline
\href{http://workshop.colips.org/dstc7/}{DSTC7-Track2}                               & Facts-grounded conversations extracted from Reddit                                                                                 \\ \hline
DSTC9 Track 1 \cite{DBLP:conf/sigdial/KimEGHLH20}                                                          & Conversations where the dialogue flow does not break when users have out of scope requests                                         \\ \hline
DSTC 8-Track 2 \cite{DBLP:journals/corr/abs-1911-06394}                                                    & Extended the DSTC 7 Track 1 by adding 3 new dimensions                                                                             \\ \hline
KdConv8 \cite{DBLP:journals/corr/abs-2004-04100}                                                           & KG-grounded multi-domain knowledge-driven conversation in Chinese                                                             \\ \hline
E2E NLG \cite{DBLP:conf/sigdial/NovikovaDR17}                                                              & A dataset in the restaurant domain                                  \\ \hline
MedDialog \cite{zeng-etal-2020-meddialog}                                                                  & Conversations in Chinese between patients and doctors covering many specialties of diseases                                           \\ \hline
Unnamed medical dataset \cite{zeng-etal-2020-meddialog}                                                    & Conversations covering many specialties of diseases                                                                                \\ \hline
EMPATHETICDIALOGUES \cite{DBLP:journals/corr/abs-1811-00207}                                               & Conversations grounded in emotional situations                                                                                     \\ \hline
\href{https://github.com/lmrojasb/benchit}{Bench’It}                                 & General knowledge evaluation for French question answering on Wikidata                                                     \\ \hline
CALOR \cite{marzinotto-etal-2018-semantic}                                                                 & Annotated French encyclopedic history texts                                                                                        \\ \hline
MovieChAtt \cite{danescu-niculescu-mizil-lee-2011-chameleons}                                              & Conversations between pairs of characters, which are matched on IMDB                                            \\ \hline
Nell \cite{NELL_paper}                                                                                    & A system that extracts information from web text to populate a KB                                     \\ \hline
\href{https://github.com/dhruvilgala/tvtropes}{TVTropes}                             & Tropes associated with examples of their occurrences in television, film, and literature                                       \\ \hline
REDIAL \cite{DBLP:conf/nips/LiKSMCP18}                                                                     & Conversations on providing movie recommendations                                                                      \\ \hline
Emotional lexicon NRC\_VAD \cite{mohammad-2018-obtaining}                                                  & Human ratings of arousal, valence, and dominance for English words                                                                 \\ \hline
Airline travel information system (ATIS) corpus \cite{DBLP:journals/taslp/MesnilDYBDHHHTY15}               & Audio and transcripts of people asking for flight information with intent categories       \\ \hline
Multimodal E-commerce Product Attribute Value Extraction \cite{zhu-etal-2020-multimodal}                   & Textual product descriptions and product images                                                                                    \\ \hline
JDDC \cite{chen-etal-2020-jddc}                                                                            & Chinese E-commerce conversation corpus with intent information                                                                     \\ \hline
E-commerce Dialogue Corpus (ECD) \cite{zhang-etal-2018-modeling}                                           & E-commerce dataset with diverse types of conversations                                                                             \\ \hline
SimpleQuestions \cite{DBLP:journals/corr/BordesUCW15}                                                      & Factoid question answering with corresponding triple fact                                                                          \\ \hline
Music Question Answering GenQA \cite{DBLP:conf/ijcai/YinJLSLL16}                                           & Open domain factoid QA                                                                                                             \\ \hline
\href{http://tcci.ccf.org.cn/conference/2014/}{NLPCC emotion classification dataset} & Emotional-annotated sentences collected from Weibo                                                                                 \\ \hline
Short-Text Conversation (STC) conversation dataset \cite{DBLP:conf/acl/ShangLL15}                          & Conversations from Weibo                                                                                                           \\ \hline
Doc2Dial \cite{feng-etal-2020-doc2dial}                                                                    & Document-grounded goal-oriented dialogues                                                                                          \\ \hline
Ubuntu Dialogue Corpus V1 \cite{DBLP:journals/corr/LowePSP15}                                              & Multi-turn dialogues                                                                                                               \\ \hline
Ubuntu Dialogue Corpus V2 \cite{DBLP:journals/dad/LowePSCLP17}                                             & An updated version of the Ubuntu Dialogue Corpus                                                                                   \\ \hline
Douban Conversation Corpus \cite{DBLP:conf/acl/WuWXZL17}                                                   & Retrieval based dataset where for each dialogue context there are multiple candidate responses                                     \\ \hline
Engaging ImageChat \cite{DBLP:conf/acl/ShusterHBW20}                                                       & Dialogues over images using many possible style trait                                                                              \\ \hline
Multi-Genre Natural Language Inference (MultiNLI) \cite{DBLP:journals/corr/WilliamsNB17}                   & Sentence pairs annotated with textual entailment information                                                                       \\ \hline
Microsoft Research Paraphrase Corpus (MRPC) \cite{dolan-brockett-2005-automatically}                       & Sentence pairs where each pair is labelled if it was paraphrased by a human                                                        \\ \hline
Multimodal Dialogue (MMD) \cite{DBLP:journals/corr/SahaKS17}                                               & Multimodal domain-aware conversations between shoppers and sales agents                                                            \\ \hline
OpenDialKG \cite{DBLP:conf/acl/MoonSKS19}                                                                  & Human-to-human utterances role-playing dialogues which are grounded with corresponding entities and paths from a KG               \\ \hline
Freebase \cite{DBLP:conf/www/BastBBH14}                                                                    & KB with data inserted mainly by its members                                                                                        \\ \hline
ATOMIC KG \cite{DBLP:conf/aaai/SapBABLRRSC19}                                                              & Commonsense reasoning KG                                                                                                           \\ \hline
DBpedia \cite{DBLP:journals/semweb/LehmannIJJKMHMK15}                                                      & Multilingual KB extracted from Wikipedia                                                                                           \\ \hline
ConceptNet \cite{DBLP:conf/aaai/SpeerCH17}                                                                 & KG has labeled edges connections between words and phrases                                                          \\ \hline
\href{https://www.wikidata.org}{Wikidata}                                            & KB extracted from Wikimedia sister projects (e.g., Wikipedia, Wikisource)                                                          \\ \hline
WordNet \cite{DBLP:conf/nips/BordesUGWY13}                                                                 & Lexical English KB which groups verbs, nouns, adjectives, and adverbs into groups of concepts                          \\ \hline
\end{tabularx}
\end{table*}

\section{Conclusion}
Knowledge helps DSs narrow the gap between human and machine communication. We survey which types of knowledge past research has added to DSs. Then, we identify and discuss existing solutions to three problems which have to be solved by knowledge-enhanced DSs: knowledge search, knowledge encoding, and knowledge incorporation. Based on theories of how our brains work, we finally propose ways to improve knowledge-enhanced DSs.

\bibliography{anthology,custom}
\bibliographystyle{acl_natbib}




\end{document}